\documentclass[3p,authoryear,preprint,review, 11pt]{elsarticle}

\usepackage[utf8x]{inputenc}
\usepackage[T1]{fontenc}
\usepackage{subcaption}

\usepackage{amsmath}
\usepackage{amssymb}
\usepackage{graphicx}
\usepackage[colorinlistoftodos]{todonotes}
\usepackage[colorlinks=true, allcolors=blue]{hyperref}
\usepackage{enumitem}
\usepackage{makecell}
\usepackage{algorithm}
\usepackage{algpseudocode}
\usepackage{fixltx2e}

\usepackage{booktabs}
\usepackage{url}

\usepackage{multicol}
\usepackage{float}

\usepackage{lineno}
\usepackage{setspace}

\usepackage{booktabs,makecell}
\usepackage[margin=10pt,
            font=normalsize,
            labelfont=bf,
            labelsep=space,
            position=below]{caption}
\journal{N/A}
\usepackage{lipsum}
\usepackage{enumitem}

\begin{document}

\begin{frontmatter}
	
	
	
	\title{Reinforcement Learning for Solving the Pricing Problem in Column Generation: Applications to Vehicle Routing}
	
	
	\author[TUE]{Abdo Abouelrous\fnref{*}}
	\ead{a.g.m.abouelrous@tue.nl}
         \fntext[*]{corresponding author}
        \author[TUE]{Laurens Bliek}
        \ead{l.bliek@tue.nl}
      \author[KU]{Adriana F. Gabor }
	\ead{adriana.gabor@ku.ac.ae}
        \author[TUE]{Yaoxin Wu}
        \ead{y.wu2@tue.nl}
        \author[TUE]{Yingqian Zhang}
	\ead{yqzhang@tue.nl}

	\address[TUE]{Department of Information Systems, Faculty of Industrial Engineering and Innovation Sciences, Technical University Eindhoven, The Netherlands}
	\address[KU]{Department of Mathematics, College of Computing and Mathematical Sciences, Khalifa University, United Arab Emirates}

\begin{abstract}
 In this paper, we address the problem of Column Generation (CG) using Reinforcement Learning (RL). Specifically, we use a RL model based on the attention-mechanism architecture to find the columns with most negative reduced cost in the Pricing Problem (PP). Unlike previous Machine Learning (ML) applications for CG, our model deploys an end-to-end mechanism as it independently solves the pricing problem without the help of any heuristic. We consider a variant of Vehicle Routing Problem (VRP) as a case study for our method. Through a set of experiments where our method is compared against a Dynamic Programming (DP)-based heuristic for solving the PP, we show that our method solves the linear relaxation up to a reasonable objective gap in significantly shorter running times.
\end{abstract}

\end{frontmatter}

\section{Introduction}

Many real-life decision-making problems such as planning, vehicle routing, scheduling problems  can be solved by employing Combinatorial Optimization (CO) techniques. In a practical setting, the focus is not on obtaining optimal solutions but rather on obtaining solutions of reasonably good quality in a short time. This is not only because optimal solutions entail long run-times which is not suitable for real-time decision-making, but also may not be so optimal when implemented in practice due to the computational approximations involved. Furthermore, optimal solutions are often subject to constant change due to changing circumstances in the operational environment \citep{horvitz2013reasoning,kwakkel2016coping, albar2009heuristics,watson2016approximate}. 

Many methods have been developed to address CO problems. These problems can often be modeled by large integer  programs and solved by  Column Generation (CG) based algorithms, such as branch and price algorithms, that solve the Linear Program (LP) relaxation at each node in a branch- and- bound procedure by a CG approach \citep{desaulniers2006column}. 

A complete explanation of CG can be found in \cite{feillet2010tutorial}. The idea behind CG is to start with a LP containing a small number of variables, called the Restricted Master Problem (RMP) and add columns with negative reduced costs successively, assuming a minimization criterion. To find the columns with negative reduced costs, a pricing problem (PP) needs to be solved repeatedly until no columns with negative reduced costs can be added. The art of designing efficient CG based algorithms is in reformulating the problem such that the Pricing problem can be solved efficiently \citep{vaclavik2018accelerating}. 
In many CO problems that are solved with CG, the PP is an Elementary Shortest Path Problem with Resource Constraints \citep{morabit2023machine}. This is a routing problem that strives to find the shortest path from a starting point to an end point without violating constraints on the resources consumed by the path such as vehicle capacity. Solving this problem exactly is a complex task as it is NP-hard \citep{dror1994note}.

Historically, the PP has been treated by Dynammic Programing (DP)-based algorithms such as labeling algorithms. For some problems instances of ESPPRC such as those encountered in difficult Vehicle Routing Problem (VRP) instances, labeling algorithms can be computationally demanding. Examples include large-scale instances or instances with wide time windows where many feasible routes exist. Furthermore, the success of these methods is mainly due to the hand-crafted configurations that may make it difficult to replicate or even generalize to other problems.

Consequently, there has been a recent initiative in the Machine Learning (ML) domain to counter the computational challenges invoked by DP-based methods. ML techniques can be seen as an alternative to making decisions in a principled way, with methods such as deep learning being well-suited to problems characterized by a high-dimensional space \citep{bengio2021machine}. These methods make use of data from the operational environment \citep{giuffrida2022optimization} by extracting information from the solution structure which heuristics do not exploit \citep{d2020learning}. This helps to define a more guided search procedure. While the solution distribution is not known, the solution generation is performed statistically with the help of mathematical optimization \citep{bengio2021machine}.

Another advantage of ML algorithms is that they can  learn from the collective expert knowledge and build on it to extend to other problems \citep{bengio2021machine}. For instance, different problems can be repeatedly solved by learning common solution structures \citep{khalil2017learning}. Although the training of ML methods – which normally happens offline is not trivial, the methods can be quickly applied to solve the problem on different instances thereafter \citep{zhang2021solving}.

\cite{mazyavkina2021reinforcement} stresses the added value of ML methods in solving CO problems with many successful applications such as in \cite{d2020learning}, \cite{hottung2020neural}, \cite{nazari2018reinforcement}, and \cite{kool2018attention}. For CG, however, ML methods have been previously used mainly for pre- and post - processing the PP and not to solve the PP itself. Most often, they use a DP to solve the PP, carrying forward the associated computational predicaments.

In this paper, we propose the first ML model that directly and independently solves the PP in CG using Reinforcement Learning (RL). We refer to our model whose function is to optimize the PP as POMO-CG. Our model iteratively constructs columns by adding one node after another until a feasible column is generated. The RL model strives to generate columns with minimum reduced cost at every CG iteration - given a minimization objective. This is in contrast to work like \cite{chi2022DRLforCG} and \cite{morabit2021machine} where the ML model deals with  selection of pre-generated columns rather than generating them itself.

Pursuing the columns with most negative reduced cost should theoretically speed up convergence \citep{lubbecke2005selected}. This pursuit has been largely obstructed by the associated computational burden it invokes as pricing problems are often NP-hard. To that end, we use ML to efficiently the search for columns with large negative reduced costs.

The paper is organized as follows. In Section \ref{Prev work} we discuss related literature and our contribution. Section \ref{RL Model} describes the general framework for training Deep Reinforcement Learning to generate new columns for the CG scheme. Section \ref{prob des} contains an application of the framework to capacitated VRP with time windows (C-VRPTW). In Section \ref{Num Exp}  we present  the set-up of the numerical experiments and discuss the obtained results. Section \ref{Critique} offers a discussion that compares our method with many other established CG methods. Finally, Section \ref{Conclusion} affirms our conclusions.

\section{Previous Work}
\label{Prev work}

Many of the PPs in CG can be solved by using Dynamic Programming (DP)-based methods. Labeling Algorithms \citep{desaulniers2010vehicle} such as the one proposed in \cite{desrochers1992new} and its many variants \citep{boland2006accelerated,chabrier2006vehicle} have been at the forefront of solving the PP in applications like VRP. For large instances, these methods often require significant storage space and computational resources. In many cases, in order to improve efficiency, the search space is restricted by means of dominance rules, which are problem specific. Furthermore,  to speed up the computation time during a branch and bound procedure, most of these methods relax some constraints at the expense of generating weaker lower bounds for the master problem,  which may result in slower convergence \citep{feillet2004exact}. 

Exact approaches to solving the PP like the ones proposed in \cite{feillet2004exact} and \cite{lozano2016exact}, strive to rectify this shortcoming. For instance, \cite{feillet2004exact}, adapts labeling algorithms to account for the elementary constraints which allow a route to visit a customer multiple times.  On the other hand, the procedure proposed in \cite{lozano2016exact} makes use of a bounding scheme to prune the solution space of the PP. However, both methods struggle with more challenging instances such as instances with large time windows or clustered customer locations. Although the algorithm proposed in \cite{lozano2016exact} is competitive in performance, it's running time may be compromised by the computation of bounds in the pre-processing step.  In order to speed up the computation, many auxiliary techniques are often integrated with this method. These methods involve tuning many parameters that can be complicated without theoretical guidelines. 

Faster local search-based heuristics to solve the PP have been proposed \citep{guerriero2019rollout}. However, these methods  are not always able to find columns with negative reduced costs, leading to premature convergence. As a result, they are often used as auxiliary techniques that support the aforementioned techniques.  For the large part, many of these methods are hand-crafted, rendering them a challenge to develop and replicate \citep{accorsi2022guidelines}.

Recently, several ML procedures have been proposed to help solve the PP. Methods like the ones described in \cite{morabit2023machine}, \cite{morabit2021machine}, \cite{chi2022DRLforCG} and \cite{xu2023enhancing} help in solving the PP by either pre- or post-processing it. Specifically, \cite{morabit2023machine} and \cite{morabit2021machine} train a model with a complex architecture based on a Graph Neural Network (GNN), which, however, has little control over the columns generated. They both make use of Supervised Learning (SL) concepts by which columns are selected after generation in the former, and arcs are selected before generation in the latter. \cite{chi2022DRLforCG}, analogously, employs a framework for column selection with RL. Lastly, \cite{xu2023enhancing} leverage RL to select a graph reduction heuristic before solving the PP with a labeling algorithm.

Since these methods  do not solve the PP directly, they use labeling algorithms which induce other computational difficulties as explained above. Additionally, these methods often employ auxiliary techniques to maximize performance. These techniques may be difficult to replicate sometimes, adding to the complexity of the method, while they do not clearly quantify the added value of the associated ML models in improving performance. That said, the contributions of this paper can be summarized as follows:
\begin{itemize}[noitemsep]
    \item We propose the first ML model that directly and independently solves the PP of ESPPRC in a CG scheme.
    \item We show how training data for dual values can be generated to train ML models for solving PPs.
    \item We illustrate the potential of our method to achieve objective values of the LP-relaxation within a reasonable gap in much shorter times compared to a DP baseline.
\end{itemize}

\section{Reinforcement Learning for Column Generation}
\label{RL Model}

In this paper we propose a framework for using Deep Reinforcement Learning (DRL) for training a model (POMO-CG) to generate new columns efficiently. The suitability of RL arises from its applicability to sequential decision-making problems that can be solved by DP and expressed as Markov Decision Processes (MDP)s. As a PP in CG can be modeled as a DP where one column component is iteratively added after another, the RL agent can decide on the column component to be added.  In the following sections, we elaborate on our RL model and its integration in CG.

\subsection{POMO-CG Framework}
\label{Combined Approach}

\begin{figure}
\centering
\includegraphics[width=15cm,height=10cm]{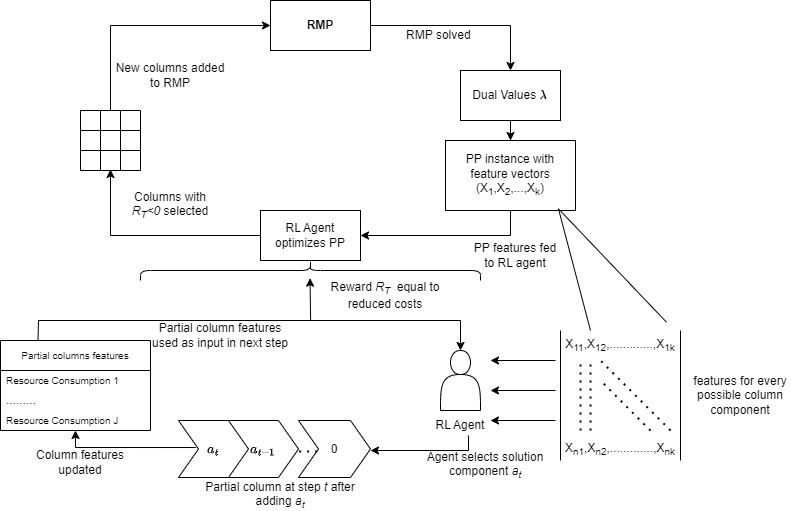}
    \caption{Visual illustration of our complete CG approach using pre-trained RL model.}
    \label{Overall Approach}
\end{figure}

The objective is for the RL model to learn to solve PPs of ESPPRC variants. It is, thus, trained on a sample of ESPPRC instances. In that sense, training is independent of the CG framework as the RL model only interacts with the PP environment without accounting for the RMP. This is in contrast to \cite{chi2022DRLforCG} where the effect on the RMP is learned during column selection which can be a complicated learning task. After training, the RL agent is integrated in CG by directly solving the resulting PP in every iteration. Figure \ref{Overall Approach} illustrates the framework by which the pre-trained RL agent is integrated in the CG loop. The PP in concern depends on our choice of problem to be solved with CG. We give an example application in Section \ref{prob des}. 

Once the master problem is solved, the dual values are extracted to formulate a PP. At each CG iteration, the RL agent iteratively solves the PP by taking into account problem features and (partial) column features. After the columns are generated by the agent, only the ones with negative reduced cost are added to the master problem. The procedure is repeated until the RL agent is unable to produce a column with negative reduced costs.

Our RL agent solves PPs independently without the interference of any heuristic. In ML terminology, such an approach is referred to as an end-to-end approach. The pseudo-code describing this approach is given in Algorithm \ref{RL for PP pseudocode}, where the RL agent is specifically referred to in line 7. Note that this is the algorithmic translation of Figure \ref{Overall Approach}. 

\begin{algorithm}
	\caption{RL for PP Optimization.}
\begin{algorithmic}[1]
 \State Input: CO problem instance, trained RL agent to solve the PP. 
 \State Let $\Omega$ be the set of generated columns, and $\omega$ the set of columns generated in current iteration.
 \State Initialize columns; start with a subset of columns that compose $\Omega$.
 \State Solve master problem with initial columns to generate duals.
 \While {columns with negative reduced costs exist and termination criteria not met.}
 \State Construct PP instance using duals from last master problem.
 \State Solve PP instance using RL agent to generate column(s).
 \If{columns with negative reduced costs are found :}
 \State{Add these columns to $\omega$.}
 \Else
 \State Stop.
 \EndIf
 \State Add $\omega$ to $\Omega$.
 \State Re-solve master problem with $\Omega$ to generate new duals.
\EndWhile 
  \end{algorithmic}
  \label{RL for PP pseudocode}
\end{algorithm}

\subsection{MDP Formulation}
\label{MDP formulation}

To define the MDP, we make use of the following notation. Let $s_t$ and $a_t$ represent the state and the action taken in step $t$. Let $S_t$ and $A_t$ denote the state and action space at step $t$, respectively. Additionally, let $r_t$ represent the reward collected after step $t$.  $s_t$ describes the current state of the problem and contains information used to decide which action $a_t$ to take. Once $a_t$ is performed, the MDP moves to state $s_{t+1}$ and reward $r_t$ is realized. The procedure is repeated a column is constructed.

For CG, the definitions of the state, action space and rewards depend on the PP being solved. In principle, the state $s_t$ describes the PP and the current state of solving the PP. More precisely, $s_t=(G_e,X_t)$,  where $G_e$ is a vector describing the static features of the problem such as parameter input that characterize the mathematical optimization model of the PP and $X_t$ is a vector containing features describing the current (partial) solution to the PP at step $t$. Similarly, action $a_t \in A_t$ corresponds to a solution component to be added to the partial solution. For instance, in a routing problem, $a_t$ can correspond to a customer (node) being added to a route at step $t$ without violating the feasibility constraints. Lastly, the reward $r_t$ typically measures the quality of the column being generated, like its reduced cost.

\subsection{Model Architecture and Training}
\label{archit}

Since many PPs can be modeled as an ESPPRC, we propose to use an architecture from DRL domain for routing. Several DRL-based architectures have been proposed for routing problems in \cite{kool2018attention} and \cite{kwon2020pomo}. Both methods leverage an attention mechanism based on a Graph Attention Network \citep{velivckovic2017graph} and make use of some benchmark algorithm known as the baseline in the loss function. \cite{kwon2020pomo}, however, adjusts the loss function to account for multiple baselines that conjointly form the shared baseline used in training. The multiple baselines are constructed from a sampling strategy by the RL agent itself, with convergence attained once each of these baselines sufficiently approximates the shared baseline in average performance.

These end-to-end methods are known to be relatively consistent in performance as instances grow in size, while repeatedly demonstrating their superiority in the routing literature. As these methods are well suited  for moderately sized VRP instances of around 100 customers like the ones we treat in this paper, we select the GNN architecture of \cite{kwon2020pomo} for our model.

The method in \cite{kwon2020pomo} employs different solution trajectories which can be very useful in a CG framework as the generation of multiple columns in one iteration is often needed to speed up convergence. Moreover, it resulted in an improved performance over a variety of routing problems compared to \cite{kool2018attention}. POMO-CG specifically makes use of an encoder-decoder architecture where the encoder treats the static features of the problem and the decoder works with dynamic features of the current solution. More precisely, the encoder generates embeddings of the graph's nodes which are then used by the decoder alongside information pertaining to the current solution to formulate the state $S_t$ described in Section \ref{MDP formulation}.

An important consideration is that the GNN architecture was originally developed for routing problems where distances between the nodes are euclidean. In PPs, however, the arc lengths are generally not euclidean with respect to the node coordinates as we will see below when discussing the VRP example. As a result, applying \cite{kwon2020pomo} to this problem would offer new insights regarding its applicability beyond the conventional scope of Euclidean problems.

Model training is done using the REINFORCE algorithm \citep{williams1992simple} which specifies a gradient loss function $\mathcal{L}(.)$. We summarize the training mechanism in Algorithm \ref{POMO-CG training pseudocode}. For a more detailed description of the training mechanism, we refer to \cite{kwon2020pomo} whose training configuration we use unless explicitly defined for some parameters in Section \ref{Num Exp}. 

\begin{algorithm}
	\caption{POMO-CG Training}
\begin{algorithmic}[1]
 \State Input: problem set $\phi$, problem size $n$, Nr. of epochs $E$, Nr. of instances per epoch $I_e$, batch size $B$, step size $\alpha$, initial model parameters $\Theta$.
 \For {each of the $E$ epochs.}
 \State Sample $I_e$ ESPPRC problems in batches of $B$ from problem set $\phi$.
 \For{each batch $B$}
 \For {each problem $b$.}
 \State{Initiate $n$ trajectories where, in each trajectory, node $i\in \mathcal{V}$ is visited after the depot.}
 \State{The RL agent generates the solution in trajectory $i$ with objective value $r^b_i$.}
\State Use average reward $\frac{1}{n}\sum_{i=1}^n r_i$ to determine baseline $\hat{r_b}$.
 \EndFor
 \State Calculate gradient $\nabla_{\theta}=\frac{1}{Bn}\sum_{b=1}^B\sum_{i=1}^n\mathcal{L}(r^b_i - \hat{r_b})$
 \State{Update model parameters: $\Theta=\Theta+\alpha\nabla_\theta$}
  \EndFor
 \EndFor

  \end{algorithmic}
  \label{POMO-CG training pseudocode}
\end{algorithm}

\section{Example Application: C-VRPTW}
\label{prob des}

To showcase the added value of the proposed ML model, we apply it to solve a common VRP variant, the Capacitated Vehicle Routing Problem with Time Windows (C-VRPTW). In this problem, vehicles start from a depot, and efficient routes need to be designed such that all customers are visited before the vehicles return to the depot. There are constraints on the vehicle capacity and on the customer time windows. Finally, there is a restriction on the time vehicles should return to the depot. 

The pricing problem for this VRP variant is known as Elementary Shortest Path Problem with Resource Constraints and Time Windows (ESPRCTW) \citep{chabrier2006vehicle}, which is a variant of ESPPRC. To the best of our knowledge, while ML methods have been previously proposed for the Traveling Salesman Problem (TSP) and VRP, there is no  ML model to solve the ESPRCTW directly. In the following sections we describe the mathematical formulation of ESPRCTW, specify the MDP framework and how POMO-CG can be used in this context.

\subsection{Mathematical Formulation}

The ESPRCTW is defined on a graph with a set of nodes $\mathcal{V}$ and a set of arcs $\mathcal{A}$. It involves finding a route of  minimal length that starts from the depot -indexed 0, visits a subset of nodes, and returns to the depot while respecting capacity and time window constraints.
Each node is located in a 2-D map, where the location is given by a vector $(x_i,y_i)$.  
Each node $i \in \mathcal{V}$ has demand $q_i$, service time $s_i$ and time window $[a_i,b_i]$. Observe that the problem size $n$ is equal to $|\mathcal{V}|$. For the depot, we have $q_0=0$, $s_0=0$, and $a_0=0$ while $b_0$ represents the operating horizon so that a route must return to the depot by $b_0$ at the latest. Given dual values $\lambda_j$ for $j \in \mathcal{V}$, the arc length is given by $p_{ij}=t_{ij}-\lambda_j$, with $t_{ij}$ being the travel time from node $i$ to $j$. Observe that $p_{ij}$ is not a euclidean distances, whereas the architecture of \cite{kwon2020pomo} was originally developed for euclidean problems with strictly positive objective functions. In contrast, $p_{ij}$ can take both positive or negative values, hence, standard shortest path algorithms cannot be used directly. By using this architecture to solve ESPRCTw, we demonstrate a new application of it. 

The following variables are needed to formulate ESPRCTW as a mathematical model
\begin{itemize}[noitemsep]
\item[-] $x_{ij}$: binary variable indicating whether arc $(i,j)$ is included in the path
\item[-] $r_i$: continuous variable representing the arrival time at node $i$
\end{itemize}

The objective function (\ref{obj}) represents the length of the path length.   Constraints (\ref{depot1}) and (\ref{depot2}) require that a route starts and ends at the depot. Constraints (\ref{flow}) ensure flow conservation at each node on the path. Constraints (\ref{subtour}) eliminate subtours. Constraints (\ref{capacity}) ensure that the route capacity is not violated. Constraints (\ref{arrival}) and (\ref{tw_con}) ensure that arrivals at any node respect the time windows. 

\begin{gather}
    \min \sum_{(i,j)\in \mathcal{A}}p_{ij}x_{ij}\label{obj}\\
    s.t. \hspace{0.5cm} \sum_{(0,j) \in \mathcal{A}}x_{0j}=1 \label{depot1}\\
    \sum_{(i,0) \in \mathcal{A}}x_{i0}=1 \label{depot2}\\
    \sum_{(i,j) \in \mathcal{A}}x_{ij}-\sum_{(j,i) \in \mathcal{A}}x_{ji}=0 \hspace{0.75cm} \forall i \in \mathcal{V}/\{0\} \label{flow}\\
    \sum_{(i,j)\in \mathcal{A}: i,j\in \overline{\mathcal{V}}}x_{ij}\leq |\overline{\mathcal{V}}|-1 \hspace{0.75 cm} \forall \overline{\mathcal{V}} \subset \mathcal{V}, 1<|\overline{\mathcal{V}}|<N \label{subtour}\\
    \sum_{(j,i) \in \mathcal{A}}q_ix_{ij}\leq Q \label{capacity}\\
    r_i+s_i+t_{ij}-M(1-x_{ij})\leq r_j \hspace{0.75cm} (i,j)\in \mathcal{A}\label{arrival}\\
    a_i\leq r_i \leq b_i \hspace{0.75 cm} \forall i \in \mathcal{V}\label{tw_con}\\
    x_{ij}\in \{0,1\} \hspace{0.15cm} \forall (i,j)\in \mathcal{A}, \hspace{0.3cm} r_i\geq0 \hspace{0.15cm} \forall i \in \mathcal{V} \label{domains}.
\end{gather}

\subsection{MDP Specification}
Due to the density of the notation used in this section, we present the following table of parameters in \ref{notation}. 

An arbitrary ESPRCTW instance is characterized by static features (information) which compose the parameter input to the problem instance. Static features directly relate to the depot and customer nodes. For ESPRCTW, the static features considered are (1) depot coordinates, (2) depot time windows, (3) node coordinates, (4) node time windows, (5) node demands, (6) node service times and (7) node dual values. These features represent a mathematical summary of the graph known as the graph embedding $G_e$ which is a linear transformation of the form $W^xx_i+b^x$ where $x_i$ is the feature vector of node $i$ and $W^x$ and $b^x$ are learnable parameters of dimension $d_h=128$ \citep{kool2018attention}.

\begin{table}
    \centering
    \small
    \begin{tabular}{|c|p{0.7\textwidth}|}
    \hline
    \textbf{Notation}& \textbf{Meaning} \\
     \hline
     $\mathcal{V}$ & set of nodes in the graph.\\
     \hline
     $\mathcal{A}$ & set of arcs in the graph. \\
     \hline
     $s_t$ &state of PP solution in step $t$.\\
     \hline
     $a_t$& action taken at step $t$.\\
     \hline
     $r_t$& reward realized at after action $a_t$ is applied.\\
     \hline
     $T$& time step at which episode ends by which all solution trajectories(see Section \ref{archit}) have generated feasible columns.\\
     \hline
     $G_e$ & matrix containing graph embeddings used as input to the RL's encoder (see Section \ref{archit}).\\
     \hline
     $y_t$& time consumed by current path at step $t$.\\
     \hline
     $q_t$& load of current path at step $t$.\\
     \hline
     $l_t$ & last node added to the current path at step $t$. Equal to $a_{t-1}$.\\
     \hline
     $m_t$ & mask signaling feasible nodes that can be visited by current path at step $t$.\\
     \hline
    \end{tabular}
    \caption{Explanation of notation used.}
    \label{notation}
\end{table}

Any solution to an ESPRCTW problem starts from a depot and visits a node thereafter - other than the depot, creating a partial path. This partial path is extended every time a node is added to it; consequently, the path's features are updated. Suppose that at each step $t-1$, a node $a_{t-1}$ is added to the partial path. Accordingly, the path's time $y_t$ is updated as per the travel time to $a_{t-1}$, waiting due to early arrival before the time window and service time at $a_{t-1}$. Similarly, the load of the path, $q_t$, is updated by the additional demand of $a_{t-1}$. The masking mechanism $m_t$ masks out nodes that can no longer be visited by the partial path after step $t-1$. Finally, $l_t$, the last visited node by the path, is updated to $a_{t-1}$. The tuple ($y_t$,$q_t$, $l_t$,$m_t$) compose the dynamic features related to the path.

A path is considered complete when the depot is added at step $t=T$, after which we have a feasible route. The length of this route, as specified in (\ref{obj}) corresponds to its reduced cost. Note that not all customers must be visited which is an important feature of ESPRCTW. The transition tuples of the MDP are defined by the following:
\begin{itemize}
    \item \textbf{State}: $s_t$  describes the partial path generated from the last node addition $a_{t-1}$. When deciding on which node to add to the path, we make use of both static and dynamic features. We refer to a state as a tuple containing ($y_t$,$q_t$,$l_t$,$m_t$,$G_e$). The state space $S_t$ is such that $y_t, q_t \in \mathbb{R}$, $l_t \in \mathcal{V}$, $m_t \in \{0,1\}^{\mathcal{V}}$ and $G_e \in \mathbb{R}^{\mathcal{V} \times d_h}$.
    \item \textbf{Action}: $a_t$ is a node that can be added to the partial path ( note that $a_t$ coincides with $l_{t+1}$ in the next step). The action space $A_t =\mathcal{V}\setminus m_t$ is the set of nodes that can be added at step $t$ without violating the constraints.
    \item \textbf{Reward}: We only consider terminal rewards $r_T$ corresponding to the reduced cost of a route  given by (\ref{obj}) and no other intermediate rewards, i.e. $r_t=0$ for $t<T$. Using the total route length as the final undiscounted reward with no intermediate rewards for routing problems is motivated in works like \cite{kool2018attention}.
\end{itemize}

\subsection{Feature Normalization}
\label{feature normalization}

Feature normalization during encoding is often necessary especially for static features as this helps generalize the models to different problem instances. We employ the scaling techniques proposed in \cite{schmitt2022learning}, where an orienteering problem similar to the VRP we address is solved. The coordinates are scaled to be in [0,1]. The time components such as time windows, service times, and travel times of the customers are scaled by the upper-bound of the the depot's time window which corresponds to the operational time horizon. The customer demand is scaled by the vehicle capacity as done in \cite{kwon2020pomo} for C-VRP. Finally, we scale the dual values by the upper-bound of the the depots time window to preserve the relative differences with travel times in ($t_{ij} - \lambda_j$).

When scaling the prices, one needs to consider that the optimal objective values of the PP converge towards 0 throughout the course of the CG procedure. From a learning perspective, this might translate to decreased rewards for certain actions and distort the learning process. To guarantee consistency in the relationship between actions and rewards, we scale each price $p_{ij}$ as follows; $p_{ij}=p_{ij}/\max(abs(\min(P)), abs(\max(P)))$ with $P$ being the matrix of prices and $abs(.)$ being the absolute value operator. The scaled values are multiplied by -1 in order to accommodate the RL convention for reward maximization.
	
\section{Numerical Experiments}
\label{Num Exp}

In this section we propose a set of numerical experiments in order to illustrate the viability of the proposed model. We train a POMO-CG model on different ESPRCTW instances with $n\in \{20,50,100\}$ nodes and then use it to solve a series of C-VRPTW instances. In accordance with previous studies on CG methods in the literature such as \cite{lozano2016exact}, we will illustrate POMO-CG by solving the root node of each C-VRPTW instance. The procedure, however, can be repeated for the rest of the nodes in the branch-and-price tree. Note that in this case, one would have to additionally configure the masking functionality in POMO-CG to account for forbidden edges assuming the standard branching scheme discussed in \cite{feillet2010tutorial}. 

In the following sections, we describe how the training and testing data for our experiments are generated, the set-up of our numerical experiments and their results. At the end of this section, we analyze the ability of the proposed ML model to generalize to instances  from a different distribution than the training data.

\subsection{Data Generation}
\label{training}

In general, attention-based models like POMO-CG are trained using randomly generated data \citep{kool2018attention,kwon2020pomo}. We generate the parameters specific to the master problem (C-VRPTW) as follows. We sample $n$ nodes uniformly at random from a square $[0,1]^2$, with $n\in \{20,50,100\}$. Customer demands are assumed uniformly distributed in [1,10]. The vehicle capacity is chosen from $\{30,40,50\}$ and increases with $n$. Service times are sampled uniformly from $[0.2,0.5]$. Travel times are represented by the euclidean distances between nodes. Lastly, for each node $v$, the lower limit $l_v$ of the time window of $v$ is sampled - as integer - uniformly from $[0,16]$, while the time window width $t_w$ is sampled from $[2,8]$. The upper limit $u_v$ is such that $u_v =\min\{l_v+t_w,u_d\}$, where $u_d$ is the depot's upper time window. We set $u_d$ to 18.

In order to generate representative dual values, one may need to know the distribution of the dual values. Such a distribution may be difficult to determine even with ample data. An alternative approach is to focus on generating representative ESPRCTW instances as opposed to realistic dual values. An  ESPRCTW instance corresponding to a PP is characterized by a set of arcs of real-valued length  $t_{ij}-\lambda_j$. Thus, we propose to generate dual values solely based on the travel times. For a number $n_c$ of customers, the dual values are positive, while they are zero for the rest. We sample $n_c$ uniformly as integer from the interval $[\frac{n}{2}, n]$. Furthermore, for each customer $j$ from the $n_c$ selected customers, we generate the  dual value $\lambda_j$  uniformly from the interval [0, $\theta \times t_j^{max}$], where  $t_j^{max}=\max_{i\in V}t_{ij}$ and $\theta$ is some scaling parameter whose choice we specify in the following section.

\subsection{Experiment Configuration}
\label{Exp Config}

For all the instances, we used $E=200$ epochs with $I_e=$10,000 episodes per epoch for training. The batch size was kept constant at $B=64$. We noticed that most of the decrease in training loss happens after approximately 50 epochs. The experiments with the CG scheme were carried out on an AMD EPYC 9654 \citep{hpc} cluster CPU node of which 100 threads were used. Training on the other hand was conducted on a GPU node with 2 Intel Xeon Platinum 8360Y \citep{hpc2} Processors and a NVIDIA A100 Acceleraor \citep{hpc3}.

Training time took approximately 1, 2, and 3.5 hours for instances of size 20, 50 and 100 respectively. Remark that these training times are very suitable for consistent training and validation. This highlights the training efficiency of the POMO-CG model compared to using the same architecture for training on problems like TSP, C-VRTPW or the Orienteering Problem~\citep{schmitt2022learning,kool2018attention,kwon2020pomo}. For ESPRCTW, the lack of the requirement that all customers must be visited results in shorter episodes and faster convergence.

For each value of $n$, training data is generated with a different value of $\theta$. We observed that the results may vary significantly with $\theta$ for different instance sizes. Therefore, we experimented with $\theta=1.1$ for $n=20$ and $\theta$ uniformly distributed in $[0.2,1.1]$ for $n=50$ and $n=100$. To determine $\theta$,  we conducted a simple grid-search over a small discrete set of lower bounds $\theta_{lb}\leq1.1$ that define the interval $\mathcal{U}[\theta_{lb}, 1.1]$ from which $\theta$ is sampled. Note that this is enabled by the short training times of POMO-CG (see Section \ref{Exp Config}). We recommend considering values of $\theta_{lb}$ that are apart at increments of 0.25 since the results do not differ significantly within this range. This would lead to the training of a maximum of 4 models before the best one is chosen.

We compare the performance of our model to a baseline that is based on an adaptation of the exact method of \cite{lozano2016exact}  where we compute the bounds on the reduced costs while solving the PP. This state-of-the art algorithm has been shown to be consistently superior among DP-based methods which are the most used class of methods in solving the pricing problem \citep{desrochers1988generalized}. The baseline initiates multiple threads from the depot to each customer and strives to find a path with negative reduced costs that ends at the depot. The nodes are visited in order of increasing arc lengths such that the nodes with smallest $p_{ij}$ values are visited first. To reduce the search space, it introduces a series of pruning strategies whereby unpromising paths are eliminated.

To prevent increasingly long-running times due to the wide-time windows of our instances from which many feasible paths can be generated, we limit the time for each thread to 10 seconds. The baseline returns the first 10 negative-reduced-cost-paths it finds within the time limit. To speed up our baseline even further, we eliminate all nodes in a PP without positive dual values $\lambda_i$ before solving \citep{barnhart1998branch}. We also use the arc reduction strategy of \cite{santini2018branch} where only the $\beta\%$ of arcs lowest $p_{ij}$ values are retained in the graph. If the baseline fails to find any negative columns, we refrain from applying this arc reduction strategy in forthcoming CG iterations. We set $\beta=0.25$ in line with the paper's recommendation.

Finally, we set the overall time limit to solving the root node with this baseline to 10 minutes. The baseline terminates either upon reaching the time limit or when it fails to find a column with negative reduced costs in a given iteration. The performance of our implementation on the well-known Solomon benchmark dataset is verified in \ref{BL verify} which asserts its reliability for comparison.

We assessed the quality of the solutions obtained by  POMO-CG based on two measures: $obj_{Gap}$ and $t_{Speed-up}$. The former refers to the average relative difference between $obj_{POMO-CG}$, the final objective value obtained by POMO-CG and $obj_{DP}$, the objective value of the DP baseline, which is calculated as follows:
\begin{equation}
    obj_{Gap}=\frac{1}{K}\sum_{k=1}^{K}\frac{(obj^k_{POMO-CG}-obj^k_{DP})}{obj^k_{DP}},
\end{equation}
where the superscript $k$ indicates the $k-$th instance. Positive values indicate that POMO-CG has a higher objective value than the baseline.
The second measure, $t_{Speed-up}$, refers to the average ratio between the time the baseline needed to reach the final objective value of POMO-CG,  $t^k_{DP}$,  and the time it took POMO-CG $t^k_{POMO-CG}$ for instance $k$, which is calculated by:
\begin{equation}
\label{tgap def}
    t_{Speed-up}=\frac{1}{K}\sum_{k=1}^K\frac{t^k_{DP}}{t^k_{POMO-CG}}.
\end{equation}
Values above 1 indicate that POMO-CG is faster than the baseline. For some instances, our RL model produces a better lower bound than the DP baseline. In such cases, $t^k_{POMO-CG}$ would correspond to the time it took POMO-CG to reach the final objective value of the baseline. We also report the number of instances $J(<)$ where this happens.

\subsection{Results}
\label{results}

For each instance size, we consider $K=50$ C-VRPTW instances whose root node we solve through the procedure described in Algorithm \ref{RL for PP pseudocode}. For both the DP baseline and POMO-CG, we initialize the columns by means of a feasible solution that greedily adds the nearest unvisited customer to the current feasible route until all customer are visited. Table \ref{experimental results 1} presents the results of our experiments for all values of $n$. For the RL agent, we also report the best configuration of $\theta$ for each $n$ in the column `$\theta$'. The relevant values of $obj_{Gap}$ and $t_{Speed-up}$ as explained above can be found in the third and fourth columns. In the third column, we show the number of instances $J(<)$ where we obtain a better lower bound. Additionally, we report the mean number of CG iterations and mean running time per CG iteration for our method and the DP baseline in the last four columns.

\begin{table}[H]
    \centering
    \begin{tabular}{cc|c|c|c|cc|cc}
    \hline
     $n$&$\theta$&\textbf{obj\textsubscript{Gap}} & \textbf{t\textsubscript{Speed-up}} & \textbf{$J(<)$}& \multicolumn{2}{c|}{\textbf{Avg. Nr. CG iter.}} & \multicolumn{2}{c}{\textbf{Avg. time $/$ iter.(s)}}\\
     &&&&$K=50$&POMO-CG&DP&POMO-CG&DP\\
    \hline 
     20 &1.1& 5.62\%&0.86&1&24&52&0.01&0.01\\
    \hline 
    50&$\mathcal{U}[0.2,1.1]$& 4.03\%&1.93&4&45&114&0.03&0.03\\
     \hline
     100&$\mathcal{U}[0.2,1.1]$&3.26\%&6.46&9&61&246&0.07&0.14\\
    \end{tabular}
    \caption{Results of POMO-CG end-to-end method compared to the DP baseline for different C-VRPTW instance sizes $n$ averaged over 50 instances for each size.}
    \label{experimental results 1}
\end{table}
For the base case with $n=20$, the RL agent does not result in a reduction in run-time with $t_{Speed-up}=0.86$, while terminating at an average gap of 5.62\% relative to the baseline. This is because these instances are rather small and can be easily solved by the baseline. Starting from $n=50$,the added value of POMO-CG becomes more evident. While the objective gap is at an average difference of 4.03\%, the $t_{Speed-up}$ ratio is 1.93 indicating that our RL agent is almost as twice as fast as the baseline. The trend continues to $n=100$ where $obj_{Gap}=3.26\%$ and the reduction in run time is almost 6.5 fold. As $n$ increases, the problems become more difficult to solve and the DP baseline struggles to scale, giving a smaller $obj_{Gap}$ and a larger $t_{Speed-up}$ ratio. 

\begin{figure}[tbp]
\begin{subfigure}{0.5\linewidth}
  \centering
\includegraphics[height=6cm, width=9cm]{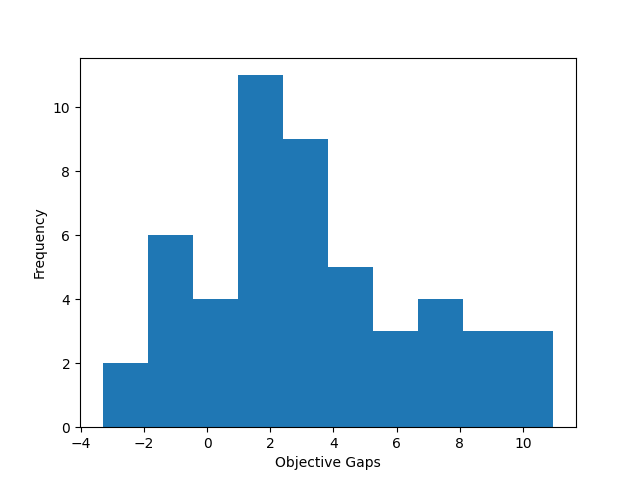}
    \caption{$obj_{Gap}$}
    \label{obj gap dist}
\end{subfigure}
\begin{subfigure}{0.5\linewidth}
\includegraphics[height=6cm, width=9cm]{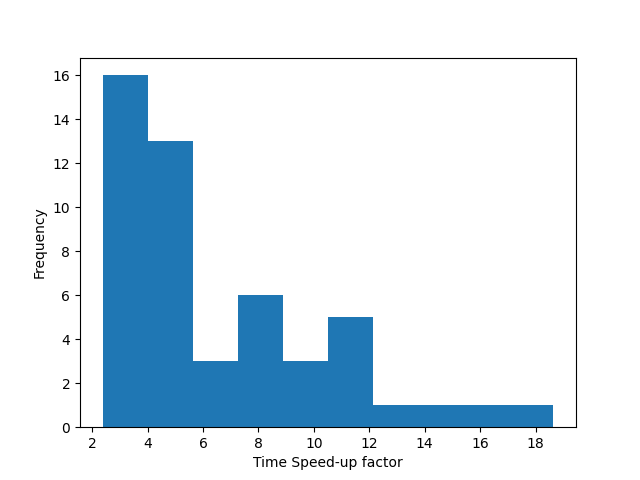}
     \caption{$t_{Speed-up}$ }
    \label{t gap dist}
\end{subfigure}
\caption{Histogram depicting the distribution of the performance measures for size size $n$=100.}
\end{figure}

To emphasize on POMO-CG's consistent ability to generate reasonably close lower bounds within a shorter computation time, we provide the histograms in Figures \ref{obj gap dist} and \ref{t gap dist} which show the distribution of the $obj_{Gap}$ and $t_{Speed-up}$ values for $n=100$. The histograms assert that the averages realized in Table \ref{experimental results 1} are due to consistent patterns among the $K$ and not outliers.

For all values of $n$, our model not only results in a smaller number of CG iterations but also in a shorter run-time per iterations. This is due to the greedy inference mechanism of POMO-CG that strives to make the best decision by only considering current information. DP methods, in contrast, iterate over the action space (i.e set of nodes that can be visited) to evaluate the impact on the overall solution, albeit at a considerable computational cost that leads to a limited improvement in the objective value. 

\begin{figure}[tbp]
\begin{subfigure}{0.5\textwidth}
  \centering
\includegraphics[height=6cm, width=9cm]{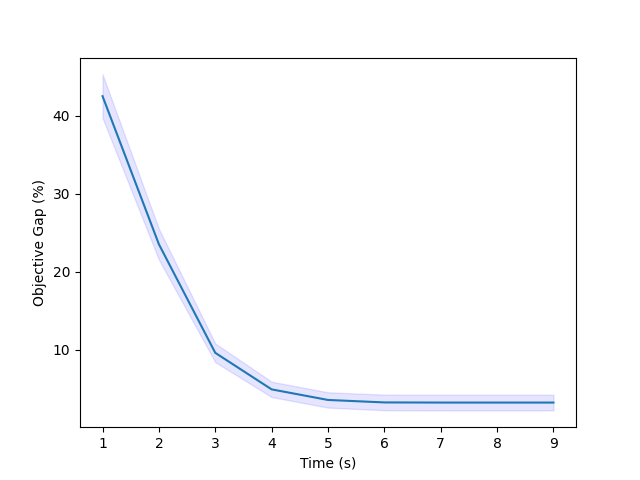}
    \caption{POMO-CG}
    \label{conv plot POMO-CG}
    \end{subfigure}%
    \begin{subfigure}{0.5\textwidth}
    \includegraphics[height=6cm, width=9cm]{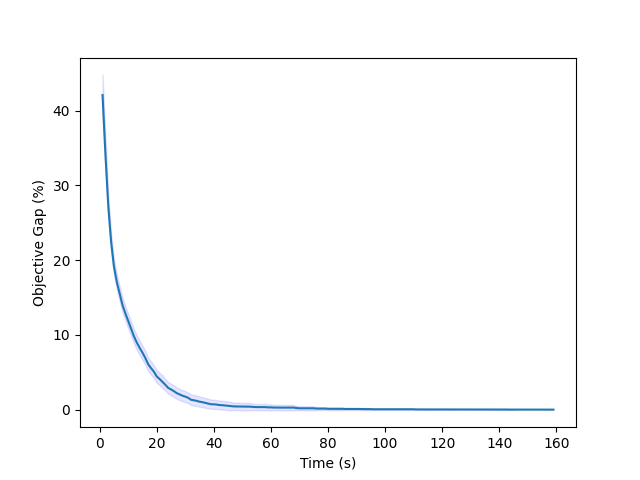}
    \caption{DP baseline}
    \label{conv plot DP}
    \end{subfigure}
    \caption{Convergence of $obj_{Gap}$ averaged over the test instances of size $n$=100. Observe the differences in run-time as indicated by the x-axis.}
    \label{}
\end{figure}

To compare the convergence of the proposed method and the baseline, we plot in Figures \ref{conv plot POMO-CG} and \ref{conv plot DP} the objective percentage gaps against time averaged over the 50 instances for $n=100$. By observing the difference in scales between the two figures, our method converges on average in less than 10 seconds,  while the baseline takes almost a minute on average. This illustrates the ability of our method to capture most of the reduction in objective value that would be attained by the baseline in a much shorter time. On the other hand, we also observe that the confidence intervals (represented by shaded areas) for $obj_{Gap}$ are slightly narrower for the DP baseline. This may be explained by the variance associated with the statistical generation of solutions with machine learning models \citep{bengio2021machine}.



\begin{figure}[tbp]
\begin{subfigure}{0.5\textwidth}
  \centering
\includegraphics[height=6cm, width=9cm]{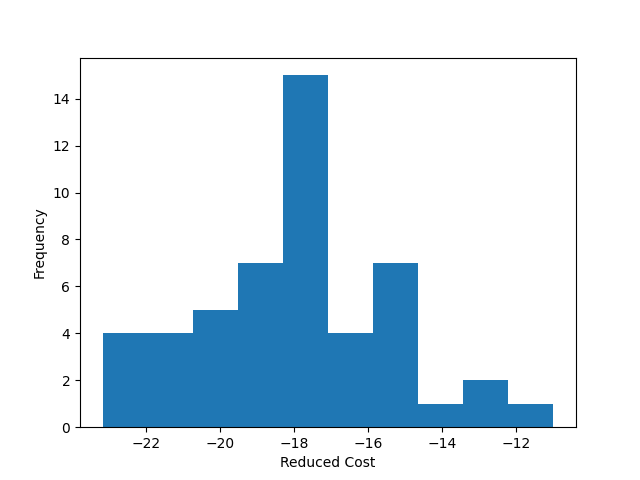}
    \caption{POMO-CG}
    \label{red cost POMO-CG}
 \end{subfigure}%
    \begin{subfigure}{0.5\textwidth}
\includegraphics[height=6cm, width=9cm]{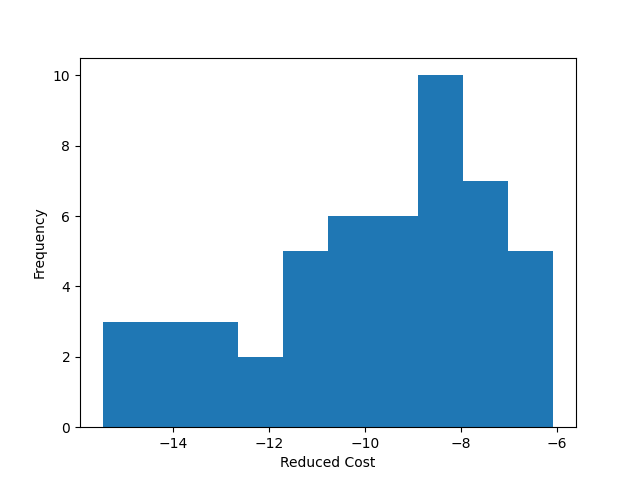}
     \caption{DP Baseline}
    \label{red cost DP}
    \end{subfigure}
\caption{Histogram depicting the reduced costs of the columns generated by both methods for the first PP in each of the test instances of size $n$=100.}
\end{figure}

To highlight the ability of our model to generate columns with larger negative reduced costs, we provide the histograms in Figures \ref{red cost POMO-CG} and \ref{red cost DP}. The histograms depict the largest reduced cost of a column generated from the first PP for each of the $50$ instances of size $n=100$. Since we initialize the CG procedure with the same set of columns for both methods, the first PPs are always the same, enabling us to directly compare the resulting reduced costs. One can observe that the reduced costs of the columns generated by POMO-CG are larger. This supports the results in Figures \ref{conv plot POMO-CG} and \ref{conv plot DP} that imply faster convergence. 

\subsection{Generalization Evaluation}
An important question concerning ML models in general is their generalization ability. ML models often fail to generalize to other problems from a different distribution in an optimization context \citep{bogyrbayeva2022learning}. Consequently, we test the model's generalization ability to get a clearer idea about its robustness and practicality. To that end, we propose two sets of experiments where, in the first set, we apply our model to solve larger instances than the ones from the training data. In the second set, we apply our model to solve a set of benchmark instances from a completely different distribution.

\subsubsection{Scalability on Larger Instances}
\label{larger instances}

In this set of experiments, we consider instances generated from the distribution described in Section \ref{training} for $n\in \{200,400,600,800\}$. These instance sizes are in line with ones reported in other studies concerned with ML application in CG such as \cite{morabit2023machine}. We set the vehicle capacities to $\{80, 100, 120, 150\}$ in increasing value of $n$. For this set of experiments, we considered $K=20$ instances for each instance size.

For all experiments, we used the POMO-CG model trained with $n=100$ and $\theta = \mathcal{U}[0.2,1.1]$, without further re-training. The results are reported in Table \ref{experimental results Large scale}. Here, the real benefit of POMO-CG  becomes more pronounced. More precisely, the pattern of decreasing objective gaps as $n$ increases becomes proceeds more dramatically. Like many DP-based method, the DP baseline struggles to scale to larger instances \citep{engineer2008shortest} given the fixed time limit of 10 minutes. On the other hand, the RL agent scales better and generates lower objective values for most of the instances as indicated by the negative values of $obj_{Gap}$ and $J(<)$.

\begin{table}[H]
   \centering
    \begin{tabular}{c|c|c|c|cc|cc}
    \hline
     $n$&\textbf{obj\textsubscript{Gap}} & \textbf{t\textsubscript{Speed-up}} & \textbf{$J(<)$}& \multicolumn{2}{c|}{\textbf{Avg. Nr. CG iter.}} & \multicolumn{2}{c}{\textbf{Avg. time $/$ iter.(s)}}\\
     &&&$K=20$&POMO-CG&DP&POMO-CG&DP\\
    \hline 
     200&3.64\%&28.59&5&101&648&0.17&0.74\\ \hline
     400&-12.71\%&35.14&20&201&721&0.35&0.77\\ \hline
     600&-16.11\%&17.40&20&301&644&0.59&0.86\\ \hline
     800&-13.39\%&9.01&20&466&527&0.96&1.06\\
     
    \end{tabular}
    \caption{Results of end-to-end approach for larger instances.}
    \label{experimental results Large scale}
\end{table}

It is only for the case of $n=200$ that the DP still manages to retain a positive $obj_{Gap}$ value and generate better lower bounds for $K-J(<)=15$ of the instances. For larger $n$, the $obj_{Gap}$ value is negative as POMO-CG finds a better objective value for all the $K=J(<)=20$ instances. Note, however, that at $n=600$, the $t_{Speed-up}$ starts to decrease relative to $n=400$ as instances become more difficult to solve even for POMO-CG. This phenomenon is even more obvious with $n=800$ as the value of $obj_{Gap}$ increases alongside the decreases in $t_{Speed-up}$ from $n=600$. This is also demonstrated by the increasing number of CG iterations and average time per iteration that approach those of the DP baseline steadily.

\subsubsection{Parameter Sensitivity}
\label{para sense}

Next we study our model's generalization ability to changes in parameter values. To do so, we consider a  new class of instances with narrower time windows and a smaller vehicle capacity. More specifically, we sample  the time window width $t_w$ from the interval $[1,2]$ and set the vehicle capacities to $\{15,20,25\}$ for $n=\{20, 50, 100\}$. For each $n$, we again consider $K=50$ instances. The distribution of the dual variables in this set of instances is different from the instances in Section \ref{training}. We use the same POMO models from Section \ref{results} for each value of $n$. The results are reported in Table \ref{experimental results tw-vary}.

\begin{table}[H]
    \centering
    \begin{tabular}{c|c|c|c|cc|cc}
    \hline
     $n$&\textbf{obj\textsubscript{Gap}} & \textbf{t\textsubscript{Speed-up}} & \textbf{$J(<)$}& \multicolumn{2}{c|}{\textbf{Avg. Nr. CG iter.}} & \multicolumn{2}{c}{\textbf{Avg. time $/$ iter.(s)}}\\
     &&&$K=50$&POMO-CG&DP&POMO-CG&DP\\
    \hline 
     20 &2.54\%&1.06&15&13&26&0.01&0.01\\
    \hline 
    50 &1.12\%&1.80&17&27&49&0.02&0.01\\
     \hline
     100&-0.62\%&5.73&29&43&98&0.03&0.05\\
    \end{tabular}
    \caption{Results of POMO-CG end-to-end method compared to the DP baseline on instances with different time window and vehicle capacity configuration.}
    \label{experimental results tw-vary}
\end{table}

The results seem to be largely in line with those observed in Table \ref{experimental results 1} as POMO-CG is able to score lower bounds within a relatively small gap in much shorter computation time. More impressively, it seems that the resulting $obj_{Gap}$ values have decreased compared to those in Table \ref{experimental results 1}. This is largely due to POMO-CG being better able to find negative-reduced-cost-routes in these constrained instances that are characterized by a smaller search space. In summary, POMO-CG does not seem to be very sensitive to changes in specific parameter values provided that the overall distribution of the instances is still similar to the training data. In the following section, we study POMO-CG's applicability to instances characterized by a completely different distribution.

\subsubsection{Different Instance Distribution}

The POMO-CG architecture was originally developed to solve a single class of instances from a predefined distribution. For distributions that are significantly different from the training data, it may fail to find columns with negative reduced costs, although they exist. In such cases, we propose a hybrid method where the PP is solved by POMO-CG until it fails to find a negative column. Thereafter, an alternative algorithm  is applied in succeeding CG iterations. This strategy has been used in previous ML applications in CG \citep{morabit2023machine}. In this context, Line 7 of Algorithm \ref{RL for PP pseudocode} would be adjusted such that the PP would be solved using the alternative algorithm should the RL agent fail. This would illustrate how our ML model is effective in solving instances from a different distribution by accelerating alternative methods. We use the DP baseline as the alternative method in our hybrid approach.

Here, we consider the popular Solomon benchmark dataset from \cite{solomon1987algorithms} with $n=100$ customers. These instances are realized from a completely different distribution of instances where the parameter values are much more variable compared to the instances from Section \ref{Num Exp}. We compare improvement in performance with the results in \ref{BL verify} with only the DP baseline. Similar to the previous section, we used the POMO-CG model trained with $n=100$ and $\theta = \mathcal{U}[0.2,1.1]$. 

The results are reported in Table \ref{experimental results Solomon 1}. The hybrid approach offers a marginal improvement in run-time compared to the DP baseline as indicated by the $t_{Speed-up}$ ratio of 1.1. Predictably, the objective values are more or less similar for all the instances as it is the DP baseline that is being deployed in later CG iterations. The reduction in run-time is much less pronounced compared to the experiments above as POMO-CG only struggles to generalize and only finds a few columns with negative reduced cost. More precisely, POMO-CG results in slightly fewer CG iterations with similar run-time per iteration.

\begin{table}[H]
    \centering
    \begin{tabular}{c|c|c|c|cc|cc}
    \hline
     $n$&\textbf{obj\textsubscript{Gap}} & \textbf{t\textsubscript{Speed-up}} & \textbf{$J(<)$}& \multicolumn{2}{c|}{\textbf{Avg. Nr. CG iter.}} & \multicolumn{2}{c}{\textbf{Avg. time $/$ iter.(s)}}\\
     &&&$K=56$&POMO-CG&DP&POMO-CG&DP\\
    \hline 
    100& 0.64\%& 1.10& 22& 530&543&0.85&0.79\\
     
    \end{tabular}
    \caption{Convergence Results of the hybrid approach against the DP baseline for each class in the Solomon benchmark instances.}
    \label{experimental results Solomon 1}
\end{table}

We noticed that POMO-CG generalizes better to instances with randomly distributed coordinates (Class $R$) than instances with clustered coordinates (Class $C$ and $RC$) as the former are more aligned with its training data. It is also noteworthy that POMO-CG does not result in computational delays. So even when applied to very different instances, in the hybrid context we can expect that there will be no significant increase in run-time. Instead, the burden lies upon providing suitable training data for generalization. These results, alongside its efficient training, serve a remarkable advantage in optimization applications.

\section{Discussion}
\label{Critique}

The focus of this study is in designing a method that generates a reasonably good objective value for CO problems within a short run-time as practical settings often entail. POMO-CG intends to quickly find as many columns with large negative reduced costs as possible. This would not only speed up convergence at every node in the branch-and-price tree, but also present a more selective criterion for adding columns to the master problem preventing it from growing too large as the branch-and-bound tree expands. Generating too many columns may be a serious issue with some popular techniques from the literature like Tabu Search that initializes the CG procedure by generating as many columns as possible, and hence, the need for a column manager like in \cite{morabit2021machine}. 

To the best of our knowledge, POMO-CG is the first ML-based framework that solves the PP without the interference of a heuristic. In the ML for CG literature, ML models are mainly employed in the pre- and post processing phase of solving the PP via labeling algorithms (see \cite{morabit2023machine} and \cite{morabit2021machine}). As a result, auxiliary algorithms are needed alongside the ML models to generate a feasible solution. While many of these auxiliary methods are considered state-of-the art, their presence induces complexity in addition to issues concerning robustness and replicability. In contrast, POMO-CG solves the PP directly, without the interference of any external technique. 

Furthermore, POMO-CG scales relatively well to larger instances when applied to ESPPRC compared to other problems like VRP as in \cite{kool2018attention} and \cite{kwon2020pomo}. This is due to the ESPPRC's property of not needing every node to be visited, reducing the number of steps needed to generate a feasible solution. One could also argue that, there is  potential for POMO-CG to generate better integer solutions as it generates lower bounds which can generate upper bounds in a less constrained fashion.

In the literature, ESPPRC is mostly solved via DP methods like labeling algorithms.  In order to render labeling algorithms efficient, one often has to reduce the state space of the DP by deriving dominance rules.  These dominance rules are problem specific and may be challenging to derive. Their derivation often requires constraint relaxation, such as allowing repeated customer visits as in the case of VRP \citep{chabrier2006vehicle}. This results in non-feasible columns being generated, which may lead to weaker lower bounds in branch and bound. DP-based methods with pruning procedures, such as \cite{lozano2016exact}, also require several auxiliary techniques and the tuning of many parameters for optimal performance.

Contrasting with the labeling algorithms, a pre-trained POMO-CG model is very fast in solving the PP and does not rely on dominance rules. Hyper-parameter tuning is also very limited as the only parameter requiring optimization is $\theta$, for which we provide guidelines. Moreover, POMO-CG only generates feasible columns which is also desirable for managerial and real-time decision-making reasons.
 
\section{Conclusion}
\label{Conclusion}

In this paper we propose a novel framework to integrate a Reinforcement Learning model in Column Generation by directly solving the Pricing Problem. We specifically consider the Elementary Shortest Path Problem with Resource Constraints (ESPPRC), which is a common Pricing Problem in many Combinatorial Optimization applications. Our model iteratively constructs routes with minimum reduced costs to speed up convergence of the Linear Program (LP) relaxation. The model is trained on a sample of ESPPRC problems from a predefined distribution. The duals are generated artificially by means of a random sampling procedure, for which we provide guidelines. The training resources needed by our model are much less than other applications using the same architecture. 

We test our method on a series of Capacitated Vehicle Routing Problems with Time Windows instances where we solve the LP relaxation and compare with another DP baseline. The results show that the proposed RL model  solves the LP relaxation within a reasonable gap and up in much shorter computation times compared to a DP baseline. We also provide a set of experimental results to demonstrate our model's generalization to larger instances and instances from a different distribution. Our model scales relatively well to larger instances, while it improves computational efficiency when combined with an alternative method to solve instances from a different distribution.

Finally, we  provide a brief discussion on the computational benefits of our method compared to existing methods from the literature and offer guidelines on the application of our method and existing methods. 

An interesting venue for future research is the generalization of the POMO-CG framework to other Combinatorial Optimization problems where the Pricing Problem is a variant of ESPPRC and explore more advanced techniques to leverage its computational value.

\section*{Acknowledgements}
Abdo Abouelrous is supported by the AI Planner of the Future programme, which is supported by the European Supply Chain Forum (ESCF), The Eindhoven Artificial Intelligence Systems Institute (EAISI), the Logistics Community Brabant (LCB) and the Department of Industrial Engineering and Innovation Sciences (IE\&IS).

\bibliographystyle{model5-names}
\bibliography{MyBIB}

\newpage
\appendix

\section{Baseline Performance}
\label{BL verify}

In this section, we validate the performance of our proposed baseline on a benchmark dataset. We opted for the Solomon dataset \citep{solomon1987algorithms} since it has often been used in the literature \citep{lozano2016exact}. The Solomon instances vary in features and test multiple aspects of C-VRPTW. In that sense, they vary in difficulty of solving with the easier being the R1, C1 and RC1 and the more challenging set being the R2, C2 and RC2. Having a baseline that delivers acceptable results on average across all instance classes would compose a reliable measure for comparing our proposed RL framework. Furthermore, the instance sizes correspond to the same one used in our training data for our largest models $n=100$. 

In Table \ref{experimental results BL verify}, we report the average gap between the objective values scored by the DP baseline and the optimal objective values reported in \cite{lozano2016exact}. We also report the average run-time until termination. The baseline gives an average optimality gap of 10.07\% with an average run-time of 461.98 seconds.

\begin{table}[H]
    \centering
    \begin{tabular}{c|c|c}
    \hline
     $n$&\textbf{Avg .Optimal Gap (\%)}& \textbf{Avg. Run Time (s)}\\
    \hline 
     100& 10.07\%& 461.98\\\hline
     
    \end{tabular}
    \caption{Results of DP baseline on Solomon benchmark instances.}
    \label{experimental results BL verify}
\end{table}

While we are aware that the optimality gaps are not very close to zero, the problem instances used in Section \ref{Num Exp} are more aligned with the class instances of R1, C1 and RC1 which are characterized by a short scheduling horizon and a limited vehicle capacity \citep{solomon1987algorithms}. For these instances our baseline scored objective values within less than 5\% of the optimality gap indicating its reliability for comparison. We must also stress that the average optimality gap of 10.06\% is not large, but rather in line with many of the heuristics used to solve the Solomon Instances in the comprehensive study of \cite{danna2005branch} for the given time limit.

\end{document}